\documentclass[a4paper, 10pt, conference]{ieeeconf}  
\IEEEoverridecommandlockouts                        
\usepackage{graphicx} 
\usepackage{siunitx}
\usepackage{amsmath}
\usepackage{subfig}
\usepackage{capt-of}
\usepackage{cite}
\usepackage{gensymb}
\usepackage[hidelinks]{hyperref}

\title{\LARGE \bf
Automated design and manufacturing of a high resolution hemispherical tactile sensor based on subtractive color mixing}
\title{\LARGE \bf Rapid manufacturing of color-based hemispherical soft tactile fingertips
}

\author{Rob B.N. Scharff$^{*1}$, Dirk-Jan Boonstra$^{*2}$, Laurence Willemet$^{2}$, 
Xi Lin$^{3}$,\\ and Micha\"el Wiertlewski$^{\dag 2}$
\thanks{* R.B.N. Scharff and D. Boonstra contributed equally to this paper}%
\thanks{$^{1}$ R.B.N. Scharff is with the Bioinspired Soft Robotics Laboratory, Istituto Italiano di Tecnologia (IIT), Via Morego 30, 16163, Genoa, Italy}%
\thanks{$^{2}$ D. Boonstra, L.Willemet, and M. Wiertlewski are with the Cognitive Robotics Department, Delft University of Technology, 2628 CD, Delft, The Netherlands}
\thanks{$^{3}$ X. Lin is with Carl Zeiss Meditec AG, Shanghai, China}
\thanks{\dag Corresponding Author (E-mail:~{\tt\small m.wiertlewski@tudelft.nl})}
}

\usepackage{color}

\usepackage[normalem]{ulem}
\newcommand{\revise}[2]{#2}

\begin{document}

\maketitle
\thispagestyle{empty}
\pagestyle{empty}

\begin{abstract}
Tactile sensing can provide access to information about the contact (i.e. slippage, surface feature, friction), which is out of reach of vision but crucial for manipulation. To access this information, a dense measurement of the deformation of soft fingertips is necessary. Recently, tactile sensors that rely on a camera looking at a deformable membrane have demonstrated that a dense measurement of the contact is possible. However, their manufacturing can be time-consuming and labor-intensive. 
Here, we show a new design method that uses multi-color additive manufacturing and silicone casting to efficiently manufacture soft marker-based tactile sensors that are able to capture with high-resolution the three-dimensional deformation field at the interface. Each marker is composed of two superimposed color filters. The subtractive color mixing encodes the normal deformation of the membrane, and the lateral deformation is found by centroid detection. With this manufacturing method, we can reach a density of 400 markers on a 21 mm radius hemisphere, allowing for regular and dense measurement of the deformation. We calibrated and validated the approach by finding the curvature of objects with a threefold increase in accuracy as compared to previous implementations. 
The results demonstrate a simple yet effective approach to manufacturing artificial fingertips for capturing a rich image of the tactile interaction at the location of contact. 
\end{abstract}

\section{Introduction} \label{sec:Introduction}

The sense of touch plays an important part of the remarkable dexterity exhibited by humans. When manipulating objects, the dense array of mechanoreceptors that innervate our skin encodes a rich representation of the contact, which include subtle cues about material properties (e.g. texture~\cite{manfredi2014natural,Wiertlewski2011}, curvature~\cite{goodwin1991tactile}, elasticity~\cite{drewing_2019}) and the dynamics of the contact (e.g. slippage~\cite{barrea2018perception}, forces and torques~\cite{johansson2004first}). In turn, these cues shape and regulate the motor commands used to robustly execute dexterous tasks~\cite{Johansson2009}. Aside from the neural inputs, the mechanical properties of the fingertip help stabilize the contact and cope with uncertainty of the object's shape and friction~\cite{warman2009fingerprints}. Similarly, soft artificial fingertips conform to the object which maximizes contact area and increases grasp stability~\cite{Fearing1988,Okamura2001}.

A large body of research has focused on realizing similar capabilities in artificial fingertips in order to aid robotic manipulation, see~\cite{Yousef2011} for an exhaustive review. Most of this research resulted in the development of soft, but flat, tactile sensors~\cite{Wang2015,Hammock2013}. However, the application of these technologies to \revise{double-curved (hemispherical) artificial fingertips}{freeform surfaces} is challenging, as the fabrication techniques~\cite{Lin2013}, or working principles~\cite{Ma2019,Yuan2015,Kappassov2019} are not easily transferred to \revise{double-curved}{freeform or even hemispherical} surfaces.

A popular approach to endow \revise{double-curved}{three-dimensional} artificial fingertips with tactile sensing capabilities is to transduce the mechanical deformation into an optical signal via a pattern of markers on the inside of the fingertip that is then captured by a camera~\cite{Ward-Cherrier2018}. This method precludes the need for direct wiring to the soft material artificial fingertip and provides high-resolution tactile information. The sensitivity and resolution of these tactile sensors is largely dependent on the quality of the signal transducing pattern on the inside of the fingertip. As most of the existing patterns fail to directly encode changes in the distance between the markers and the camera, the normal deformation is commonly approximated from the lateral deformation of the markers.

\begin{figure}[!t]
   \centering
   \includegraphics[width=\columnwidth]{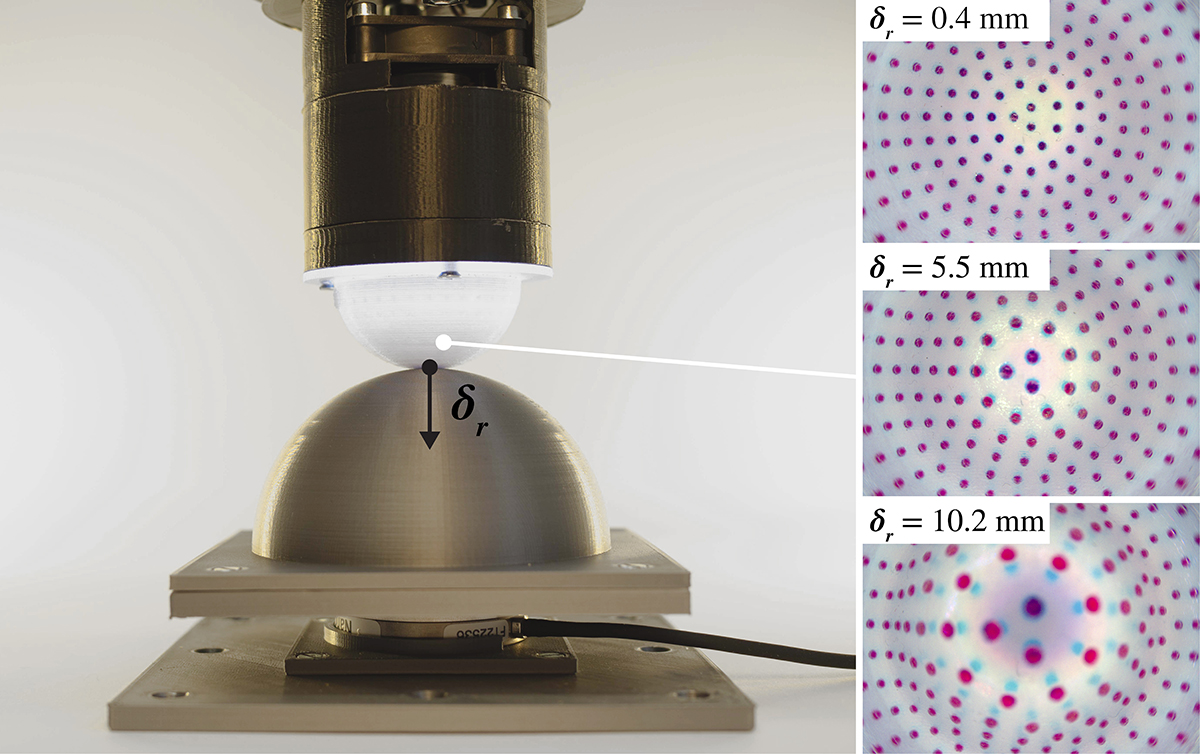}
   \caption{The 3D-printed ChromaTouch tactile sensor. Normal and lateral forces are transduced into changes in hue, centroid and apparent size of the markers.}

\label{fig:teaser}
\end{figure}

The ChromaTouch sensor was developed to more accurately capture the normal deformation by encoding it in the hue-value of the markers~\cite{Lin2019,Lin2020}. This change in color is achieved through the change in distance between two sets of overlapping optical filters of different colors. The technique was effectively demonstrated on both a flat and hemispherical surface. \revise{However, the fabrication of the hemispherical version of the sensor is laborious. After embedding the two layers of overlapping markers in a flat layer of silicone, the flat projection was folded onto a rigid hemisphere. The distortion from the projection itself as well as from the stretching that was required to fold the thick silicone layer onto the hemisphere led to a poor alignment between the two layers of markers and a non-uniform distribution of the markers across the hemisphere surface. Moreover, the fabrication technique severely limits the marker density and minimum fingertip size. Finally, the rigid core limited the maximum indentation of the tactile sensor to $1.4~\mathrm{mm}$. Hereby, the contact surface and thus number of markers within the contact surface is typically small, resulting in a relatively low accuracy for tasks such as curvature detection.}{}

In this work, we present an improved version of the ChromaTouch sensor with precise marker alignment, high marker density, and a more uniform distribution of the markers across the hemisphere. The new ChromaTouch tactile sensor (see Fig.~\ref{fig:teaser}) is capable of indentations up to $10~\mathrm{mm}$. The hemispherical transducer is fabricated using PolyJet additive manufacturing. An automated design tool was developed to easily customize the design in terms of marker size, marker density, initial distance between the overlapping markers, and fingertip size. As the selection of these parameters is no longer limited by the design and fabrication techniques, they can be selected such that a desired sensor performance is obtained.

\section{Related Work}

\subsection{RGB-based sensing of soft material deformation}
RGB cameras are a popular choice for sensing soft material deformation as they are readily available, and the high resolution provided by the camera allows for the capturing of numerous degrees of freedom. Several methods have been proposed to exploit the RGB-information provided by the camera to enhance sensor performance. Gelsight uses red, green, and blue LEDs to illuminate the surface of interest from different directions~\cite{Yuan2015}. The RGB channels of the camera are used to extract a separate image of the surface for each of the lighting directions from a single RGB image. Photometric stereo technique is then applied to estimate the surface normals from these three images. Colored light from different directions also forms the basis of the sensing principle of the flat tactile sensor by Kappassov et al.~\cite{Kappassov2019} and the finger-shaped GelTip sensor~\cite{Gomes2020}. 

Alternatively, white light can be used to illuminate the soft material, where colored elements embedded in the soft material transduce deformation to changes in the color of the light that is reflected back to a camera or color sensor.
Bai et al. apply differently colored regions inside a stretchable optical fiber to identify the location along the fiber at which deformation occurs~\cite{Bai2020}. Scharff et al. use a structure of magenta colored elements that are initially occluded by cyan colored elements. Deformation of the soft material structure leads to the appearance of the magenta elements and thus results in a change of color measured by a color sensor~\cite{Scharff2018}. The color information was shown to offer a significant increase in sensor accuracy as compared to only using light intensity information~\cite{Scharff2019}. 
Rather than using layers of occluding elements, GelForce uses two layers of differently colored markers that can both be observed by the camera~\cite{Sato2010}. The main advantage of this approach is that the shape of the markers on the outer layer remains detectable, allowing for precise tracking of the marker centroids to calculate the lateral deformation. However, the marker density of the GelForce sensor is kept low to prevent overlap between markers when a lateral deformation is applied to the sensor. This poses a constraint on the tactile resolution that can be obtained. ChromaTouch addresses this issue through the use of subtractive color mixing (see Fig.~\ref{fig:teaser}). The color of the translucent markers on the inner layer is mixed with the color of the opaque markers behind them.
As a result of the translucency of the markers on the inner layer, the centroids of the markers on the outer layer can still be tracked while the markers overlap. Consequently, a higher number of markers can be realized onto the surface. The use of colored markers instead of black and white markers was shown to drastically increase the robustness of the sensor to pixel density and lighting conditions~\cite{Lin2019}.

\subsection{Manufacturing of soft hemispherical transducers}
\revise{}{The fabrication of the previous version of the ChromaTouch sensor was laborious~\cite{Lin2020}. After embedding the two layers of overlapping markers in a flat layer of silicone, the flat projection was folded onto a rigid hemisphere. The distortion from the projection itself as well as from the stretching that was required to fold the thick silicone layer onto the hemisphere led to a poor alignment between the two layers of markers and a non-uniform distribution of the markers across the hemisphere surface. Moreover, the fabrication technique severely limits the marker density and minimum fingertip size. Finally, the rigid core limited the maximum indentation of the tactile sensor to $1.4~\mathrm{mm}$. Hereby, the contact surface and thus number of markers within the contact surface is typically small, resulting in a relatively low accuracy for tasks such as curvature detection. Alternative fabrication strategies for fabricating hemispherical transducers include silkscreen printmaking~\cite{Sato2010}, direct silicone casting of the hemisphere with markers~\cite{Winstone2013}, silicone casting of the hemisphere with the markers pasted onto the inner mold~\cite{Sakuma2018}, and multi-material additive manufacturing~\cite{Ward-Cherrier2017}.}

Strategies for distributing the markers across the hemispherical surface include a rotationally symmetric layout~\cite{Ward-Cherrier2017}, and geodesic dome representations~\cite{Chorley2009}.

However, the rotationally symmetric layout has a highly non-uniform distribution of markers, resulting in a non-uniform resolution across the hemispherical surface. On the other hand, the use of a geodesic polyhedron poses significant modeling challenges and cannot be generated for an arbitrary number of markers.

In this work, \revise{we propose a tactile sensor with an arbitrary number of equi-distributed overlapping markers fabricated through multi-material additive manufacturing.}{we present an approach for realizing a high resolution ChromaTouch tactile sensor using an automated design tool and multi-material additive manufacturing. The approach is demonstrated on a hemispherical ChromaTouch sensor with 200 markers, but can be applied to freeform surfaces with arbitrary numbers of markers as well.}

\section{Tactile sensor design rationale} \label{sec:Study}
\subsection{Subtracting color-mixing under external loads}
\begin{figure}[!h]
   \centering
   \includegraphics[width=\columnwidth]{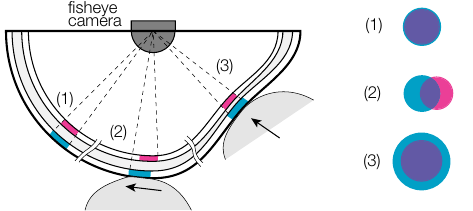}
   \caption{The hemispherical soft fingertip comprising two rows of color filters deforms under shear (2) and compressive forces (3). The membrane deformation depends on the depth, such that each row of filters is affected in a distinct manner. The optical signature of each marker can be directly linked to the displacement of the marker with respect to the camera.}
\label{fig:principle}
\end{figure}

The fundamental principle of how the sensor is able to resolve the three-dimensional displacement at the surface is illustrated in Fig.~\ref{fig:principle}. Initially (1), the sensor is not deformed, and the markers appear as a constant color that is a blend of the cyan and magenta layers. Because of elasticity, when shear forces are applied (2), the cyan layer, which is closer to the surface, moves tangentially with larger amplitude than the magenta layer. Therefore, the sub-image formed by the two layers is now the superposition of two circles, with only the intersection being subjected to the subtractive mixing. If only a normal force is applied (3), the elastomer between the two layers is compressed and the cyan layer appears as a larger portion of the sub-image.

The pattern over each sub-image is unique to the deformation that occurs at the interface, and therefore we show in the last section of this paper that it is possible to find a transformation that maps the sub-images to the actual deformation of the cyan markers, which are closer to the surface and therefore less sensitive to stress filtering caused by elasticity~\cite{shimojo1997mechanical}.

\subsection{Marker density affects measurement uncertainty}
Additive manufacturing allows us to easily vary the density of markers, and we can ask ourselves what is the influence of the marker density. Per sampling theory, the higher the density, the smaller the features of the contact that can be resolved. With tightly packed small markers, each crevasse and asperity down to the size of two markers can be registered, according to Nyquist theorem. But at the same time, the displacement of smaller markers is also more difficult to resolve by the camera, and leads to a degraded signal-to-noise ratio.  

One of the advantages of ChromaTouch over simply counting the number of pixels in black and white markers, is that the color provides a decent signal even if only 16 pixels are recruited \cite{Lin2019}. Therefore, we can freely increase the density, thus providing a good sensitivity to small features, while keeping the signal-to-noise ratio to a reasonable level. 

Round markers were chosen such that the increase in surface area resulting from normal forces is independent of the direction of the lateral forces.

\subsection{Elasticity and force sensitivity}

The signal-to-noise ratio also benefits from having a soft material construction. A given interaction force $F$ would produce a larger deformation of the membrane and therefore a more visible change of color. Considering a simple elastic model, we can write the displacement $\delta$ as:
\begin{equation}
   \delta = \frac{F\,t}{A\,E}
\end{equation}
where $t$ is the thickness of the domes and silicone assembly, $A$ the area of contact and $E$ the equivalent Young’s modulus of the multi-layer assembly. This equation illustrates well that in order to have a large sensitivity to small forces, the Young’s modulus must be as small as possible. Similar reasoning can be made in shear, where the shear modulus, which is typically three times smaller than the Young’s modulus, should also be small for producing large and noticeable tangential displacements. 
\revise{The Polyjet additive manufacturing method allows us a fine control over the elasticity of the printed domes. In this example, they are made out of  Agilus, Shore hardness A30, which provides a desirable high compliance. However, the compliance also comes with long viscoelastic relaxation time. We circumvent this issue by adding an elastic silicone compound in between the domes, which shows very little viscous effect, and decreases the overall dynamic response of the membrane.}{The final 40~mm-diameter dome deformed 3.5~mm under 10~N of normal force.}

\section{Design and Manufacturing} \label{sec:Design}
\revise{The automated design and manufacturing process allows for rapid customization of the tactile sensor through adjusting parameters such as hemisphere diameter, marker size, and number of markers. A family of ChromaTouch sensors is shown in Fig.~\ref{fig:ChromaTouchFamily}.}{} The design and manufacturing process \revise{}{of the ChromaTouch sensor} is discussed in this section.

\subsection{Parametric design tool}
\subsubsection{Marker distribution}
The chosen strategy for creating a uniform marker distribution on a hemispherical surface for an arbitrary number of points is described below. Points on a spherical surface are created using a spherical coordinate system, where $\theta \in[0,\pi]$ is the polar angle, $\varphi \in[0,2\pi]$ is the azimuthal angle, and $R$ is the radius of the outer hemisphere.

\begin{figure}[h]
	\centering
	\includegraphics[]{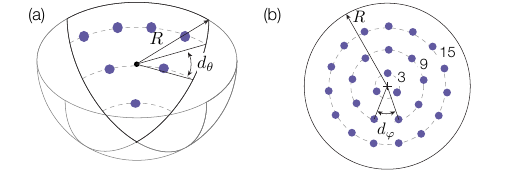}
	\caption{(a) Isometric view of the marker placements over one quarter of one of the hemispheres. (b) Top view of the distribution of markers.}
	\label{fig:grasshopper1}
\end{figure}

 The average area per point is equal to $2 \pi R^2/N$, where $N$ is the number of points to be generated on the hemisphere. Circles of latitude are created at constant intervals $d_\theta$. Then, points are placed on these circles with a distance $d_\varphi$ from each other, where $d_\theta\simeq d_\varphi$, and $d_\theta d_\varphi$ is equal to the average area per point (see Fig.~\ref{fig:grasshopper1}). The detailed algorithm is described by Deserno~\cite{Deserno2004}. This algorithm was implemented in Grasshopper~\cite{Grasshopper2021} to generate $2N$ uniformly distributed points on a sphere. These points form the centers of the bases of $2N$ cones with their base normal to the surface of the sphere, their apex at the centroid of the sphere, and a cone radius equal to the markers. Finally, the inner and outer set of markers are defined as the intersection between these cones and the inner and outer domes respectively.

\subsection{Multi-material additive manufacturing}
\begin{figure}[!ht]
   \centering
   \includegraphics[width=\columnwidth]{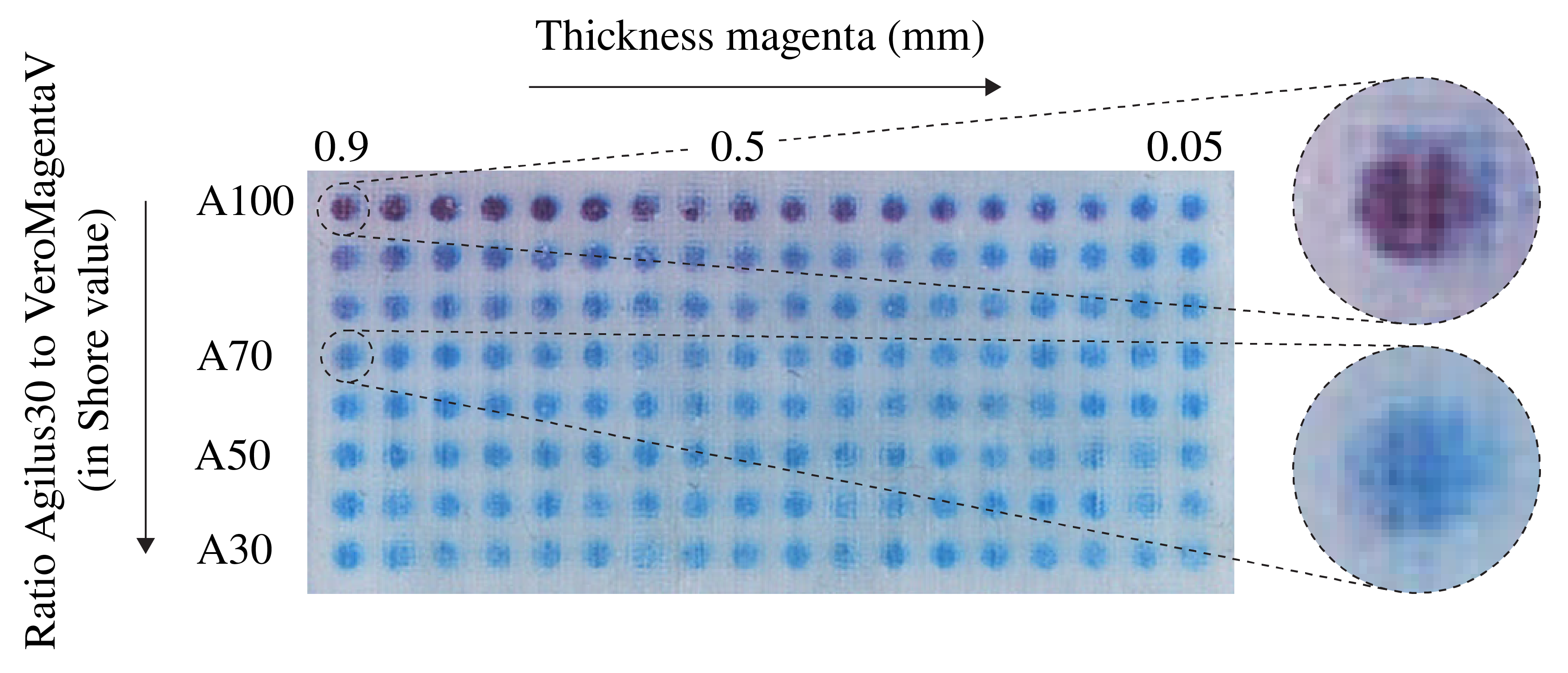}
   \caption{Influence of the marker thickness and material mixture on the color mixing. The thickness and material mixture of the cyan markers on the bottom layer is kept constant ($1~\mathrm{mm}$, VeroCyanV, Shore value A100). The thickness of the magenta markers influences their transparency dramatically. The insets show two combinations. The top inset is the optimal as it shows a good color mixing.}
\label{fig:colormixing}
\end{figure}

The domes with the colored markers are manufactured on a Stratasys J735 multi-material additive manufacturing system. The PolyJet technology allows for the combination of up to seven different photopolymers in a single printed object. The colors of the markers are chosen from the translucent VeroVivid family (i.e. VeroCyanV, VeroMagentaV, and VeroYellowV). The markers on the outer hemisphere were printed in VeroCyanV, as this was found to be the least translucent material in the VeroVivid Family. The mixing of the colors of the inner and outer markers can be tuned through the choice of marker color for the inner hemisphere, the thickness of the markers, and digital mixing of the colored material with a flexible transparent material. As the mechanical performance of the domes benefits from a minimal thickness of the more rigid markers, the more opaque VeroMagentaV was selected over the more translucent VeroYellowV for the markers on the inner dome. Fig.~\ref{fig:colormixing} shows the subtractive color mixing of the cyan and magenta markers.

The thickness of the inner VeroMagentaV markers range from $0.9~\mathrm{mm}$ to $0.05~\mathrm{mm}$ in steps of $0.05~\mathrm{mm}$ along the x-axis, whereas the digital mixture of the colored markers with Agilus30 is varied along the y-axis according to Stratasys' predefined material profiles (Shore A values from top to bottom: $100$, $95$, $85$, $70$, $60$, $50$, $40$, $30$). Using a white background behind the outer cyan markers and a light source with a similar intensity as that in the final prototype, the best results were obtained using a marker thickness of $0.9~\mathrm{mm}$ and a Shore A value of 100. The domes are printed in the flexible transparent Agilus30, with a Shore hardness of 30A. \revise{}{However, the compliance also comes with long viscoelastic relaxation time. We circumvent this issue by adding an elastic silicone compound in between the domes, which shows very little viscous effect, and decreases the overall dynamic response of the membrane.} The rigid frame that is used to connect the domes to the camera is printed using the rigid VeroClear material. The domes are printed separately to \revise{ease the removal of}{circumvent the need for removing} support material in between the domes.

\subsection{Postprocessing and assembly}
\begin{figure}[!ht]
   \centering
   \includegraphics[]{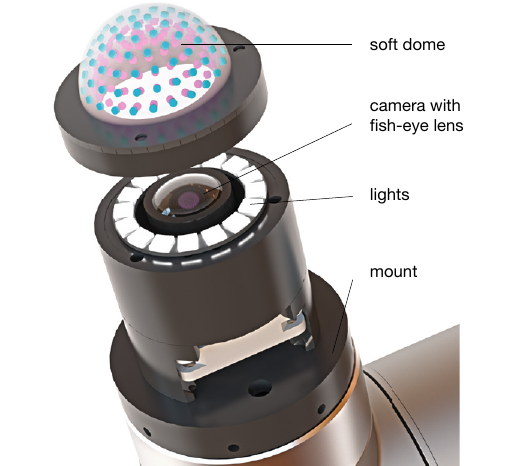}
   \caption{Render of the fingertip assembly with the tactile sensing dome, fish-eye camera and LED lights, reprinted from~\cite{boonstra2022}.}
\label{fig:sensor_assembly_render}
\end{figure}

As shown in Figs.~\ref{fig:sensor_assembly_render} and \ref{fig:ChromaTouchFamily}, the inner and outer domes are printed together with a rigid frame. These frames simplify the assembly and align the marker arrays. Two holes in the rigid frame enable the casting of silicone in between the domes. A liquid uncured silicone (Smooth-On SORTA-Clear 12) is injected between the two domes with a syringe. The excess air escapes through the second hole, avoiding bubble formation. During the casting process, the dome assembly is placed within a mold to prevent deformations in the hemisphere, and left to cure. Once the two domes are bonded together, a $1~\mathrm{mm}$ thick layer of white pigmented silicone is cast on the outside of the dome. This diffusive layer has two purposes: it blocks light coming from external disturbances, and it is used to diffuse the sensor's own light along the inside of the dome. An embedded USB-camera (Basler Dart daA1600-60uc with a Basler Evetar M118B029520IR fisheye lens attached) tracks the markers on the surface of the domes. The optical deformation of the $178$ degree viewing angle projects all markers from the hemisphere onto the imager with an equal area projection. Placing the focal point at the center of the hemisphere minimizes additional digital preprocessing of the images. An Adafruit NeoPixel Ring 16 provides the internal lighting. The 16 RGB LEDs are bright enough to overcome potential residual disturbances, and drive the camera at its maximum frequency with short exposure times. The on-board LED drivers allows them to be easily controlled with an Arduino Uno. The entire assembly of dome, camera, lens and lighting is supported by a 3D-printed enclosure, as can be seen in Fig.~\ref{fig:sensor_assembly_render}.

\section{Measurement setup} \label{sec:MeasSetup}
\subsection{Data acquisition}
In order to compare the performance of the new sensor to that of the previous generation of the ChromaTouch, the same curvature estimation strategy as described in~\cite{Lin2020} was followed. The sensor assembly was mounted on a Universal Robot UR5 robotic arm for precise and repetitive manipulation. Three 3D-printed objects with radii $-40$, $\infty$, and $40~\mathrm{mm}$ are mounted on a rigid base. The robot pushed the artificial fingertip into the objects with indentations ranging from $0$ to $10~\mathrm{mm}$. The robot moved step-by-step, 1~mm at the time. Between each step, the robot paused to make sure no viscoelastic effect is measured, and a deformation image is taken. To minimize the effects of friction, the objects are lubricated with Cyclon dry weather lube between every two trials. A single trial comprises 12 images at different indentations and one baseline image without deformation. Total data collection consists of six trials for all three curved objects, resulting in a relatively small dataset of 234 images.

\subsection{Signal processing}
The images obtained from the camera with the fisheye lens are processed as follows. First, a Gaussian filter is applied, and to mitigate the effect of non-uniform lighting, the background is subtracted. Once the correction is done, the images are converted from RGB to HSV color space. The hue channel is used to segment the markers using a threshold that lies between the cyan and magenta hue. The centroid of each cyan marker is detected from the binary images using the {\it regionprop} function in Matlab.

\begin{figure}[!ht]
    \centering
   \includegraphics[width=.9\columnwidth]{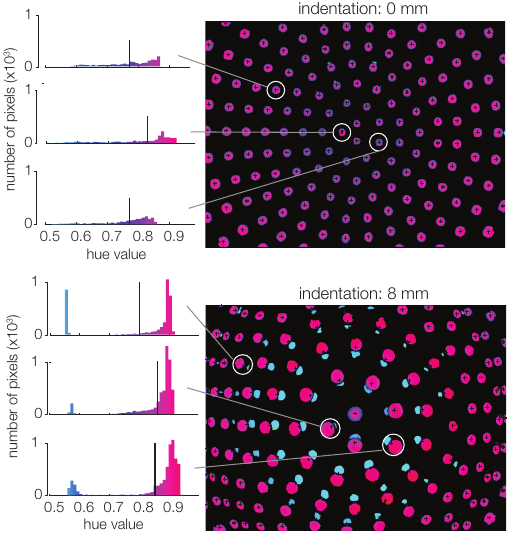}
   \caption{Histogram of the hue channel of three markers in the HSV colorspace at 0 and 8~mm of normal indentation. The vertical line indicates the average hue value.}
\label{fig:histograms}
\end{figure}

At this stage, we have the hue values of the pixels of each marker, which encode the normal displacement \revise{}{(Fig.~\ref{fig:histograms})}, and the motion of the centroids in two dimensions, which encodes the lateral displacement of the markers.

\section{Validation and results} \label{sec:Results}

\subsection{Calibration of the sub-image pattern to displacement}

After preprocessing, the images need to be calibrated to make precise curvature estimations. The exact color of each marker can be slightly influenced by construction, illumination and camera parameters. Therefore, we proceed to find a transform that robustly converts the sub-image around each marker into the effective displacement. To find this transformation, we calibrate our sensor by measuring a series of indentations on curved and flat objects and calibrating it against a ground-truth analytical model.

\begin{figure}[!ht]
   \centering
   \includegraphics[width=\columnwidth]{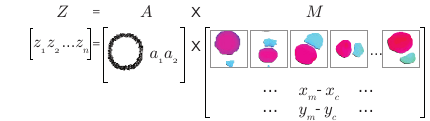}
   \caption{Computation of the mapping between every marker sub-image $M_i$ and their location ($x_m, y_m$) to the measured displacement $z_i$.}
\label{fig:calibration}
\end{figure}
The ground truth model is based on the Hertzian contact theory, which predicts a parabolic displacement at the surface of the membrane. To capture the diffusion of stresses at the depth of the marker, caused by the elasticity of the medium, we filter the parabolic shape using Boussinesq-Cerruti equations. This model shows that the theoretical normal displacement at $2~\mathrm{mm}$ (the thickness of the silicone layer in between the cyan and magenta markers) below the surface can be well approximated by a Gaussian function such that:
\begin{equation}
    u_\mathrm{z}(r)|_{z=-2\mathrm{mm}}=\delta \exp\left( -\frac{r^2}{a^2}\right)
\end{equation}
where $a$ is the contact area between the two spherical objects and $\delta$ the relative displacement of the bodies. 

In practice, we first extract $101 \times 101$-pixel windows \revise{}{corresponding to $2.8 \times 2.8$~mm} around each marker for different indentations and for different object curvatures. These sub-images were reshaped into a column vector, see Fig.~\ref{fig:calibration}. We append to this vector the global centroid of the marker, then each vector forms the columns of a matrix $M$. In addition, we computed the theoretical normal displacement $z_i$ from the ground-truth model and formed the output vector $Z$. 
The calibration matrix $A$ is found by solving: 
\begin{equation}
A = Z \cdot M^+
\end{equation} with $M^+$ the Moore-Penrose pseudo-inverse of the data matrix $M$. By the mere presence of images from all conditions, $A$ is invariant to the curvature of the contacting object and the normal indentation of the robot.

\subsection{Curvature estimation}

\begin{figure}[!ht]
   \centering
   \includegraphics{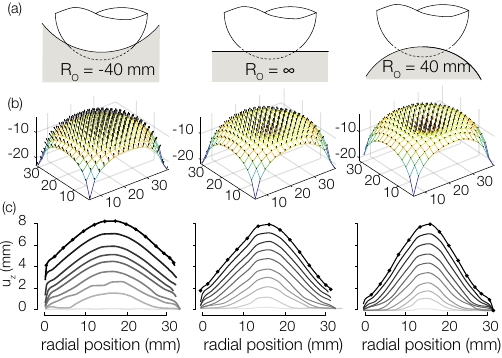}
   \caption{(a) Sensor pressing on a negative, null and positive curvature object. (b) The resulting shape after an indentation of 8~mm. (c) Profile of the displacement of the central cross-section of the sensor. Curves show the progression from 1 to 8~mm indentation.}
\label{fig:contactComparison}
\end{figure}

Using the calibration matrix, we were able to reconstruct the normal displacement field experienced by the sensor from the observed images sampled at the location of the markers. Fig.~\ref{fig:contactComparison}(b) shows the interpolated displacement on a regularly spaced grid for the contact with the flat and the two curved objects of radii $\pm 40~\mathrm{mm}$. The resulting profile of the displacement of the middle cross-section are shown in Fig.~\ref{fig:contactComparison}(c) along with the result of the curve-fitting with the filtered Hertzian contact model.

To estimate the radius of curvature from the 3D displacement, we are assuming that the object is infinitely rigid compared to the sensor and that the friction can be neglected due to the presence of lubricant. 
The theoretical values for the equivalent radius of curvature $R_{\mathrm{eq}}=(R_\mathrm{s}^{-1}+R_\mathrm{o}^{-1})^{-1}$, with $R_\mathrm{s}=21~\mathrm{mm}$ and $R_\mathrm{o}=\{-40, \infty, 40\}~\mathrm{mm}$ indicating the radii of the hemispherical sensor and the objects respectively, can be calculated as $44.2~\mathrm{mm}$, $21.0~\mathrm{mm}$, and $13.8~\mathrm{mm}$  for the negative, null, and positive curvature object respectively. The equivalent radius can be estimated such that:
\begin{equation}
    R_{\mathrm{eq}} = \frac{a^2}{\delta}
\end{equation}

\begin{figure}[!ht]
   \centering
   \includegraphics[width=\columnwidth]{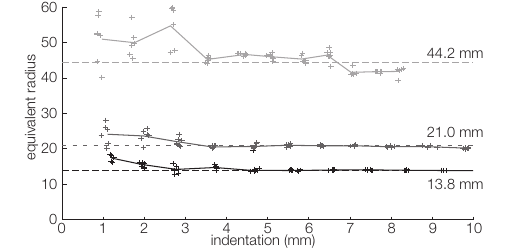}
   \caption{Results of the curvature estimation for an increasing normal indentation. Found values approach for larger indentations the theoretical values depicted by the dashed lines.}
\label{fig:radiusComparison}
\end{figure}

Fig.~\ref{fig:radiusComparison} shows the estimation of the equivalent radius for an increasing normal indentation and for the 6 trials. For indentations lower than one millimeter, the noise of measurement has a significant influence on the quality of the results. However, after a $1~\mathrm{mm}$-indentation, the estimation of the equivalent radius converges to the true value. Due to the increased compliance, the deformation was more apparent than the version presented in \cite{Lin2020}. The increase in deformation, together with a more accurate manufacturing, led to a threefold increase of the accuracy at the highest compression possible. 

\section{Discussion and Conclusion} \label{sec:Conclusions}

In this work, we present an automated design and manufacturing method for a high resolution hemispherical ChromaTouch tactile sensor that is based on subtractive color mixing. It is demonstrated that the novel hemispherical ChromaTouch sensor significantly outperforms the previous version in curvature estimation tasks. This improvement can be attributed to the higher marker density and the capability to explore objects with much larger indentations. Hereby, the contact surface area and the number of markers within this area are increased significantly. The precise fabrication and alignment of the markers have significantly reduced nonlinear responses of the transducer, allowing for a straightforward calibration process that only requires a small dataset.

\begin{figure}[!ht]
   \centering
   \includegraphics[width=\columnwidth]{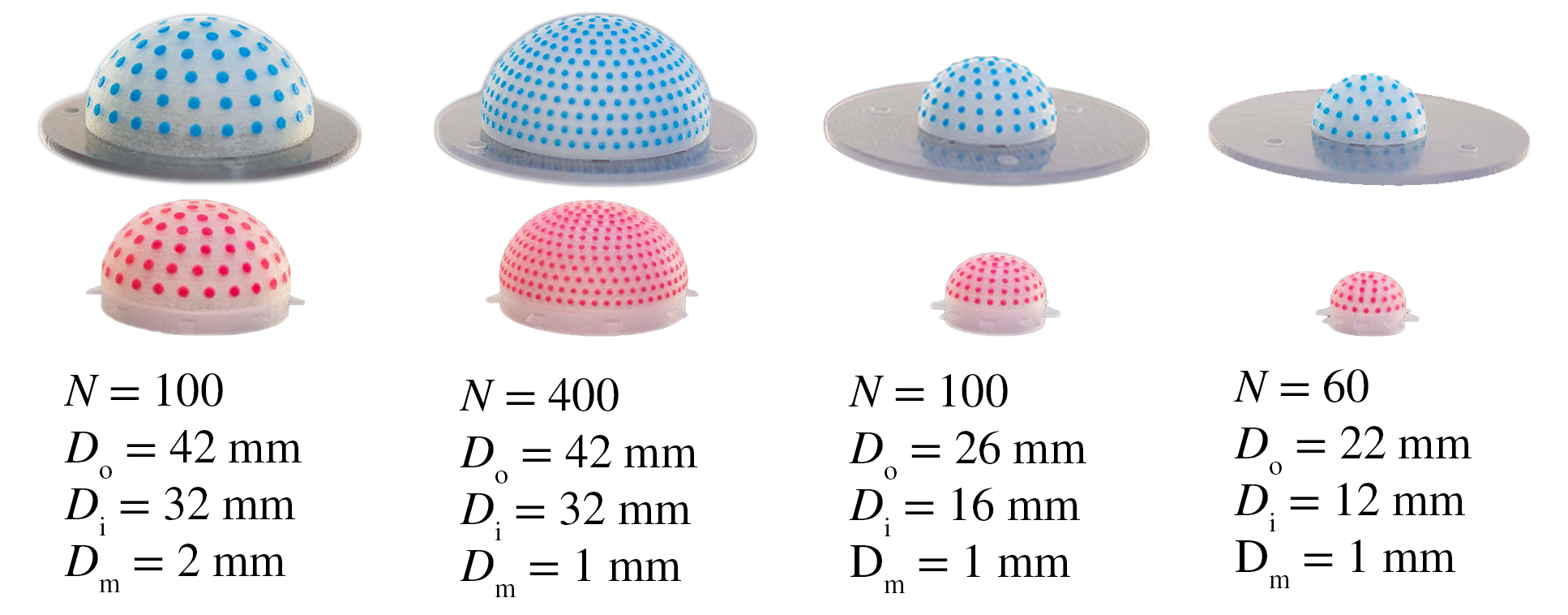}
   \caption{The ChromaTouch family: The automated design and manufacturing process allows for rapid customization of the tactile sensor through adjusting parameters such as the inner diameter $D_i$, outer diameter $D_o$  (including the $1~\mathrm{mm}$ thick diffusive layer), marker diameter $D_m$, and marker number $N$. All 3D-printed dome layers have a thickness of $1~\mathrm{mm}$.}
\label{fig:ChromaTouchFamily}
\end{figure}

\revise{}{The automated design and manufacturing process allows for rapid customization of the tactile sensor through adjusting parameters such as hemisphere diameter, marker size, and number of markers. A family of ChromaTouch sensors is shown in Fig.~\ref{fig:ChromaTouchFamily}.}

Future work will focus on increasing the accuracy of the contact estimation through employing a convolutional neural network to calibrate the sensor.

A limitation of the new manufacturing technique for the ChromaTouch tactile sensor is that all the available colored photopolymer resins are rigid. Therefore, in contrast to the previous versions of ChromaTouch, the strain of the markers on the outer layer upon normal deformation is limited, and the color change is dependent only on the decreasing distance between the two layers of markers. The development of stretchable and colored photopolymer resins for the PolyJet Additive Manufacturing system would solve this issue. Alternatively, future work will investigate the embedding of small 3D-printed structures in between the layers of markers that act as mechanical levers to amplify the color change upon normal deformation. Such structures would bypass the need for new resins. 

In this work, the automated design and manufacturing process is demonstrated for hemispherical sensors of various sizes, marker numbers, and marker sizes. However, the approach is not limited to a hemispherical shape and can be extended to freeform surfaces. Future work will investigate the integration of the transducers in freeform surfaces in order to realize tactile sensing capabilities in a wide range of products (e.g. a steering wheel).

\section*{ACKNOWLEDGMENT}
\revise{}{We thank TU Delft students Tiemen Kos, Mathijs de Heer, Mattia Strocchi, Luca Elbracht, and Mark Kruijthoff for their dedication to the development of early prototypes of 3D printed tactile sensors. We also thank Willemijn Elkhuizen for her precious insights. This work has been supported by the 4TU Soft Robotics program.}


\bibliographystyle{IEEEtran}
\bibliography{BIB.bib}

\end{document}